\documentclass[
]{ceurart}

\usepackage{todonotes}
\usepackage{listings}
\usepackage{tabularx}
\usepackage{booktabs}
\usepackage{makecell}
\usepackage{comment}
\usepackage{amsmath}
\usepackage{amssymb}
\usepackage{amsfonts}
\usepackage{svg}
\usepackage{tikz}
\usetikzlibrary{arrows.meta,positioning,fit,backgrounds}
\usepackage{pifont}

\begin{document}

\copyrightyear{2026}
\copyrightclause{Copyright for this paper by its authors.
  Use permitted under Creative Commons License Attribution 4.0
  International (CC BY 4.0).}


\title{PyKEEN-NSX: A Modular Framework for Static, Dynamic and Schema-Aware Negative Sampling in PyKEEN}


\author[1]{Ivan Diliso}[%
orcid=0009-0007-2942-202X,
email=i.diliso1@phd.uniba.it,
]

\author[1]{Nicola Fanizzi}[%
orcid=0000-0001-5319-7933,
email=nicola.fanizzi@uniba.it,
]

\author[1]{Claudia d'Amato}[%
orcid=0000-0002-3385-987X,
email=claudia.damato@uniba.it,
 ]

\address[1]{Dipartimento di Informatica, University of Bari Aldo Moro, Bari, Italy}

\conference{}


\begin{abstract}
\textit{Embedding methods} have become popular due to their scalability on link prediction and/or triple classification tasks on Knowledge Graphs (KGs). Embedding models are trained relying on both positive and negative samples of triples. However, since KGs generally contain only positive assertions, negative samples are 
artificially generated through negative sampling strategies, ranging from simple random corruption to more sophisticated approaches that exploit structural, semantic, or embedding information. The design and implementation of advanced negative samplers remains challenging, as most popular Knowledge Graph Embedding (KGE) libraries provide support only for basic strategies and lack a unified framework for developing more advanced and customized solutions.
To address this gap, we introduce PyKEEN-NSX, an extension of PyKEEN, the popular KGE framework, that provides a modular engineered abstraction for negative sampling.
The proposed architecture separates the generation of candidate negative pools, conditioned on an explicit context, from the selection strategy, enabling the development and integration of static, schema-aware and dynamic approaches within a consistent framework. 
Based on this abstraction, we implement six negative samplers, while remaining fully compatible with existing PyKEEN workflows and pipelines. 
As a proof of concept, we study negative availability across four datasets, showing that  constrained pools frequently fall below the requested number of negatives, so that the encoded criterion is to a large extent replaced by the random fallback that supplements them.
\end{abstract}


\maketitle

\section{Introduction}
\label{sec:introduction}

\textit{Knowledge Graphs} (KGs) represent data in a graph structure as factual statements, possibly enriched with schema-level knowledge, in the form of triples (subject, predicate, object), and are effectively used in numerous knowledge-intensive applications~\cite{gashteovski2020aligning,li2023survey}.
Because of their inherently distributed nature, KGs remain incomplete, which has motivated (automated) completion tasks and the development of \textit{Knowledge Graph Embedding} (KGE) models, largely used for this purpose. KGE encodes entities and relations in a low-dimensional vector space.
Training such models relies on a contrastive approach, in which observed triples must be distinguished from 
\textit{negative samples}~\cite{madushanka2024negative}.
Since KGs consist mostly exclusively of positive statements, negatives are artificially generated, 
under the \textit{local closed-world assumption} (LCWA)~\cite{Nickel16}, by  corrupting randomly observed triples, i.e.\ by replacing their subject or object with an entity sampled from the KG.
Random corruption, however, may yield trivial or false negatives, and low-quality negatives lead to suboptimal embeddings and degraded performance on downstream tasks such as \textit{link prediction}.
Following the foundational work on KGE models~\cite{bordes2013translating}, several negative sampling approaches have been proposed:  exploiting entity similarity and relational semantics~\cite{kotnis2017analysis,dash2019distributional}, network structure~\cite{wang2022leveraging}, type constraints~\cite{krompass2015type}, the distribution of entities over relations~\cite{wang2014knowledge}, and adversarial objectives~\cite{cai2018kbgan,zhang2019nscaching}.
Despite this variety, and despite the value of schema-level knowledge for deriving explicit negative statements~\cite{jain2021improving}, KGE libraries support only basic strategies, while advanced ones remain scattered across separate repositories, often tightly coupled to the embedding method they were introduced with.
This is in part a consequence of how such frameworks are structured: negative sampling is exposed as a single operation, so every strategy must reimplement batching, target selection, filtering and tensor handling to express the criterion that distinguishes it. The interplay between representation model and sampling strategy is thus impractical to investigate, and comparing samplers on equal terms becomes difficult.
This work starts from the observation that the diversity of negative samplers is largely confined to a single component: any strategy factors into the generation of a pool of candidate entities and the selection of negatives from it, differing only in the former.
Targeting \textit{PyKEEN}~\cite{ali2021pykeen}, one of the most popular libraries for building KGE models, we introduce PyKEEN-NSX, a modular extension that makes this factorization explicit: the components common to all samplers are provided once, and a new strategy is obtained by defining its pool alone, while remaining usable with every existing KGE model available in the framework.
Six samplers, spanning static, schema-aware and dynamic corruption, are implemented on this basis.
As a proof of concept, we quantify \emph{negative availability} across three datasets, showing that constrained pools, frequently fall below the required number of negatives, so that the intended criterion is to a large extent replaced by a fallback mechanism, e.g. random corruption, without being fully aware of it.
In the following, Sect.~\ref{sec:basics} presents the abstraction and instantiates it over the literature, Sect.~\ref{sec:resource-description} describes the resource, and Sect.~\ref{sec:exp} reports the analysis. An extended version of this work is available as a preprint~\cite{d2025enhancing}.

\section{A General Abstraction for Negative Samplers}\label{sec:basics}
Negative samplers are usually presented as a heterogeneous catalogue (random, structural, schema-aware, adversarial) and implemented as self-contained, mutually incompatible components, yet they admit a common decomposition. 
Corrupting a positive triple $\tau = (h,r,t)$ of a KG $\mathcal{K} = \langle \mathcal{T}, \mathcal{A} \rangle$, with assertions $\mathcal{A}$ and schema $\mathcal{T}$ when available, replaces its head, tail, or both with entities drawn from the entity set $\mathcal{E}$~\cite{chen2023negative}. 
Given a corruption target $s \in \{\mathit{head}, \mathit{tail}\}$, any sampler factors into: a \textbf{pool generator} $\mathcal{P}_s(\tau; \Omega) \subseteq \mathcal{E}$, returning the entities admissible as corruptions of $\tau$ on $s$ under a context $\Omega$, i.e.\ the assertions $\mathcal{A}$, the schema $\mathcal{T}$, or the state $\theta$ of an auxiliary model; and a \textbf{selector} $\sigma(\mathcal{P}_s(\tau; \Omega), k) \in \mathcal{E}^k$, drawing the $k$ negatives used for training.
The factorization instantiates over the literature. Among \emph{static} strategies, whose context is fixed and whose pool is therefore precomputed once, \emph{Random}~\cite{bordes2013translating} and \emph{Bernoulli}~\cite{wang2014knowledge} take $\mathcal{P} = \mathcal{E}$ and require no context, the latter drawing the corruption target $s$ with a relation-dependent probability; \emph{Corrupt}~\cite{socher2013reasoning} ($\Omega = \mathcal{A}$) restricts the pool to the entities observed in position $s$ for relation $r$; \emph{Typed}~\cite{krompass2015type} ($\Omega = \mathcal{T}$) to those satisfying the domain/range of $r$ or sharing a class with the corrupted argument; and \emph{Relational}~\cite{kotnis2017analysis} ($\Omega = \mathcal{A}$) to those linked to the fixed argument by some relation $r' \neq r$. \emph{Dynamic} strategies take $\Omega = \theta$ and must rebuild the pool per batch: \emph{Nearest Neighbor}~\cite{kotnis2017analysis} takes the $k$ entities closest to the corrupted argument under the auxiliary embedding, and \emph{Adversarial}~\cite{kotnis2017analysis} takes the $k$ closest to the prediction of $\theta$ for target $s$. All of the above leave $\sigma$ as a uniform draw. The remaining variation lies in $\sigma$: \emph{NSCaching}~\cite{zhang2019nscaching} draws uniformly from a per-triple cache of high-scoring negatives refreshed by importance sampling, \emph{self-adversarial} sampling~\cite{sun2019rotate} keeps $\mathcal{P} = \mathcal{E}$ but weights the draw by the softmax of the model score, and \emph{KBGAN}~\cite{cai2018kbgan} samples from a generator, trained by policy gradient, over a random candidate subset of $\mathcal{E}$.
Two observations follow. First, variation concentrates in $\mathcal{P}$: every strategy that does not consult a score at draw time leaves $\sigma$ as a uniform draw, so a new sampler is fully specified by its pool. Second, the static/dynamic distinction is not a taxonomic divide but a property of $\Omega$: pools conditioned on $\mathcal{A}$ or $\mathcal{T}$, both fixed, are precomputed once, whereas those conditioned on $\theta$ must be rebuilt per batch as training proceeds. Crucially, once $\mathcal{P}$\ is explicit, its size becomes measurable, turning a usually implicit assumption into a verifiable one: a strategy is meaningful only where $|\mathcal{P}_s(\tau; \Omega)| \geq k$, a condition constrained pools frequently violate (see Sect.~\ref{sec:exp}).

\section{PyKEEN-NSX}\label{sec:resource-description} 

\begin{figure}
    \centering
    \includegraphics[width=0.8\linewidth]{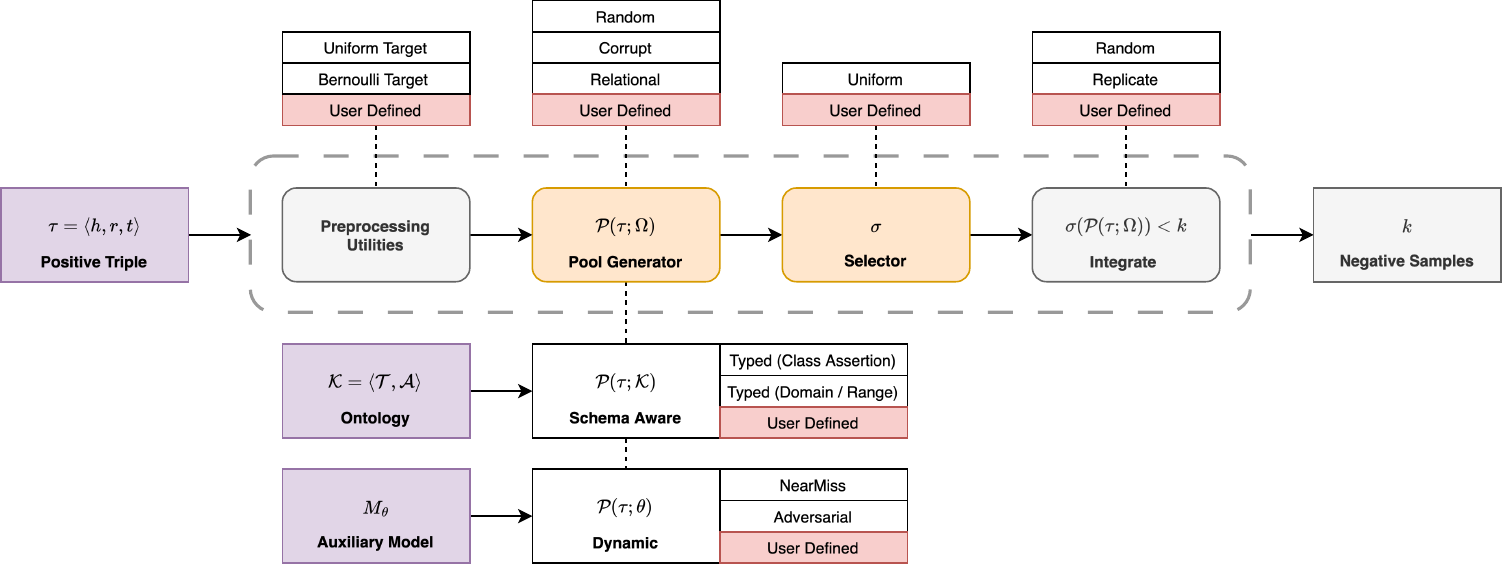}
    \caption{Architecture of PyKEEN-NSX. The abstract base class provides the shared corruption utilities and a default uniform selector $\sigma$; a new strategy is obtained by defining its pool generator $\mathcal{P}$ through the highlighted hooks. Custom classes introduced by the extension are shown under/over each component.}
    \label{fig:nsx}
\end{figure}

PyKEEN-NSX realizes the abstraction of Sect.~\ref{sec:basics} as an extension of \textit{PyKEEN}. The resource is built around an abstract class that inherits from \textit{PyKEEN}'s \texttt{NegativeSampler} and provides, once, the utilities every sampler would otherwise reimplement: replication of the positive batch, choice of the corruption target, tensor assembly, caching, and the fallback policy, applied when a pool is smaller than requested, in agreement with the literature.
The two components of the abstraction, the pool generator $\mathcal{P}$\ and the selector $\sigma$, are exposed as hooks (Fig.~\ref{fig:nsx}). A uniform selector is provided by default, so a new strategy is obtained by defining $\mathcal{P}$\ alone; the same holds for the remaining components, whose default behavior may be redefined, down to the batch-level corruption routine itself.
On this basis, PyKEEN-NSX ships six samplers. The static \emph{Corrupt}, \emph{Relational} and schema-aware \emph{Typed} (in a domain/range and and entity-class variant) precompute their pools, the latter from an OWL ontology ingested by the extension's preprocessing utilities; the dynamic \emph{NearestNeighbour} and \emph{Adversarial} condition their pools on the state $\theta$ of an auxiliary model and rebuild them per batch.
The latter accept any pre-trained model exposing \textit{PyKEEN}'s \texttt{ERModel} interface, together with a user-supplied prediction function, keeping the auxiliary model decoupled from the sampling logic. Since the base class conforms to \textit{PyKEEN}'s sampler interface, all six are usable throughout its training, evaluation and hyper-parameter optimization pipelines without modification.
Making $\mathcal{P}$ an explicit object has two practical consequences beyond ease of extension. First, the fallback that supplements an undersized pool, usually hard-coded, becomes an exposed parameter (\texttt{integrate}): when a pool yields fewer than $k$ candidates, the sampler can draw with replacement, top it up with randomly generated entities, or return the undersized pool as is. This makes the point at which a strategy degrades into random corruption both controllable and measurable directly from the interface without any training. Second, standardizing the pool under one base class makes samplers directly comparable; to our knowledge, no other KGE library offers these strategies pre-implemented under a common interface.
Documentation\footnote{https://ara-t3.github.io/pykeen-nsx/}, worked examples for training and standalone analysis, and consistent formatting accompany the code\footnote{https://github.com/ara-t3/pykeen-nsx} to support reuse.



\section{Preliminary Negative Sampling Analysis}\label{sec:exp}

\begin{figure}
    \centering
    \includegraphics[width=\linewidth]{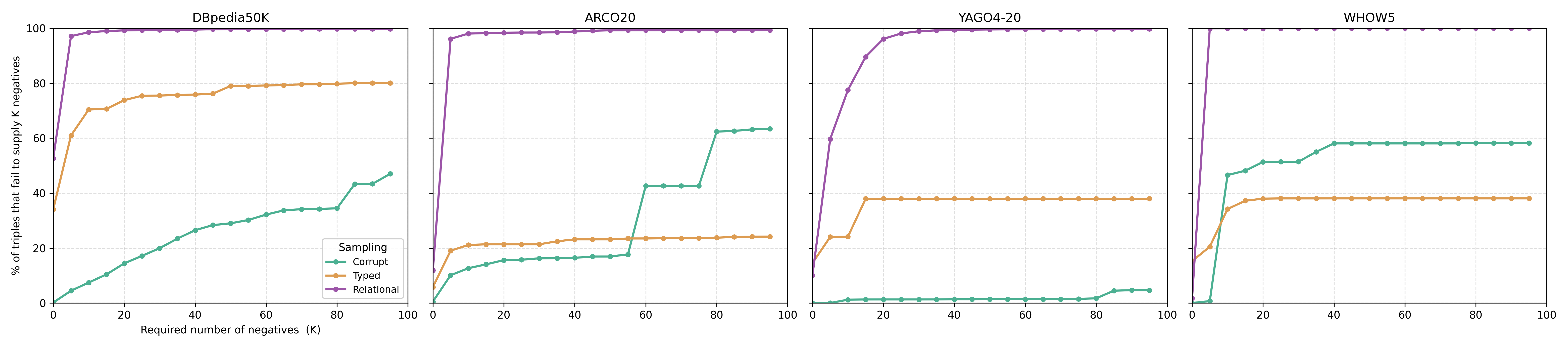}
    \caption{Negative availability: for each $k$, the percentage of training triples whose pool cannot supply $k$ negatives, and on which the sampler therefore falls back on random corruption ($k=0$ counts empty pools).}
\label{fig:pools}
    \label{fig:exp}
   
\end{figure}

The number of negatives $k$ drawn per positive is a standard parameter of KGE training. Under random corruption it is unproblematic: the pool is the whole entity set, so any $k$ is satisfied. Once $\mathcal{P}$ is constrained by structural or semantic criteria this guarantee is lost, and the pool varies from triple to triple. Where it is too small, the common remedy is to supplement it with random entities~\cite{kotnis2017analysis}, which secures $k$ at the cost of overriding the criterion the strategy encodes. The extent of this substitution is rarely reported, yet where most triples admit no pool of the requested size, the training signal is determined largely by random corruption.
Since PyKEEN-NSX exposes $\mathcal{P}$, this can be measured through the interface itself, without training. Fig.~\ref{fig:pools} reports, for each sampler and dataset, the fraction of triples whose pool falls below $k$, i.e.\ the share of the training set on which the strategy must fall back on random corruption. The analysis covers four datasets: \texttt{YAGO4-20}~\cite{pellissier2020yago}, \texttt{DBpedia50K}~\cite{lehmann2015dbpedia}, \texttt{ARCO20}~\cite{diliso2026returnschemabuildingcomplete} and \texttt{WHOW5}~\cite{diliso2026returnschemabuildingcomplete}. Schema-aware strategies require a $\mathcal{T}$ that standard KGE benchmarks do not distribute; we therefore rely on the ontology-equipped datasets provided by~\cite{diliso2026returnschemabuildingcomplete}. Dynamic samplers are excluded, since their pool is fixed by construction at the $k$ nearest entities, hence a parameter of the strategy rather than a property of the data.
Fig.~\ref{fig:exp} compares the samplers on link prediction with RotatE on ARCO20, the dataset richest in schema axioms, under default \textit{PyKEEN} hyperparameters and $k=40$, a common high-negative setting, each in its pure form and with random integration. The two figures agree: the gain from integration follows the shortfall of Fig.~\ref{fig:pools}, and the integrated score approaches that of \emph{Random} ($0.686$) as the shortfall grows. \emph{Relational}, which cannot supply forty negatives for $99\%$ of ARCO20 triples, recovers $0.674$ and is thus indistinguishable from random corruption, whereas \emph{Corrupt}, short on $16\%$, reaches only $0.370$. Dynamic samplers are unaffected, their pools being fixed at $k$ by construction. We emphasize how the shortfall curve acts as a compatibility check between strategy and dataset, computable before any training: it gives
the largest $k$ at which a strategy still is itself, and so whether it is applicable at all on a given KG.

\begin{figure}
    \centering
    \includegraphics[width=0.6\linewidth]{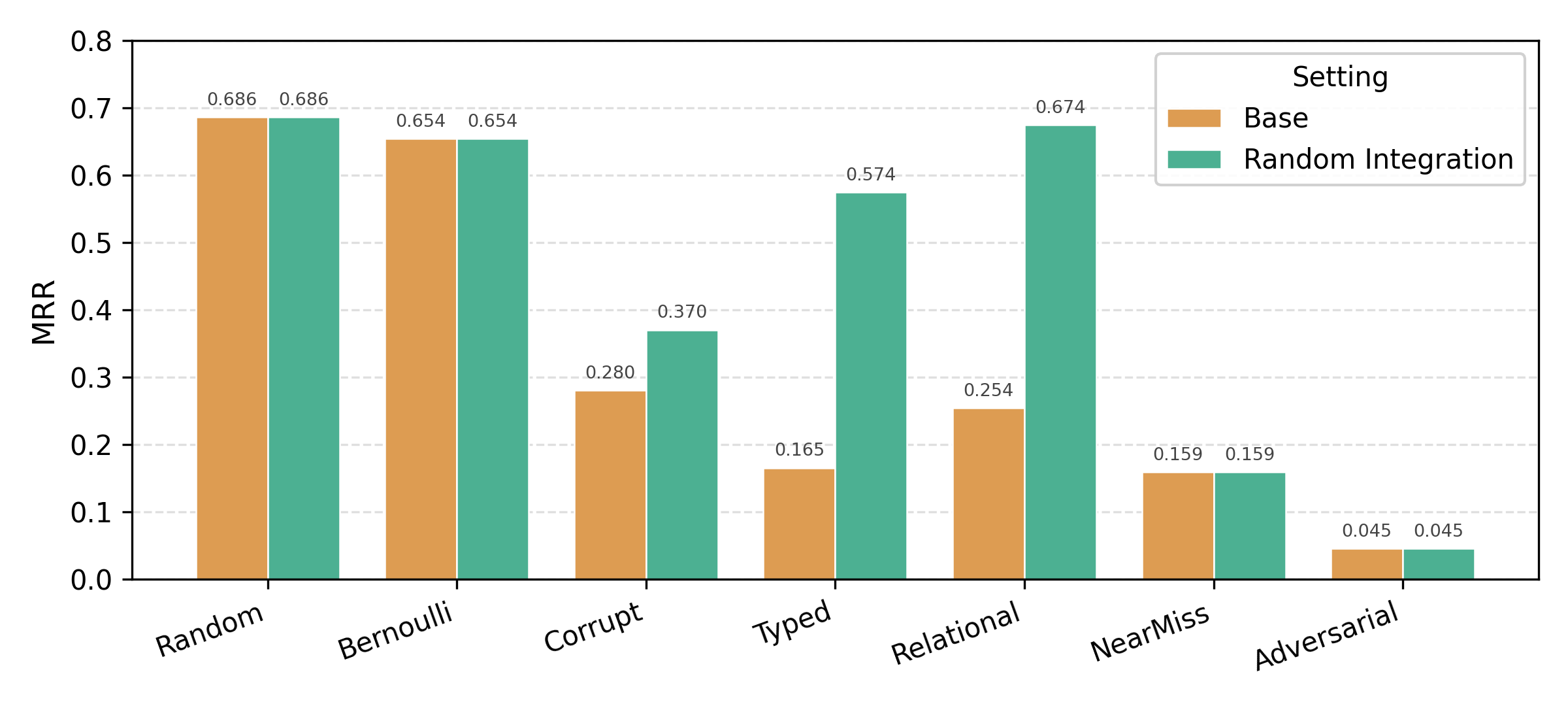}
    \caption{Link prediction evaluation (Mean Reciprocal Rank) on ARCO20 (RotatE, $k=40$), for each sampler in its pure form and with random fallback (\texttt{integrate}).}
    \label{fig:exp}
   
\end{figure}

\section{Conclusions and Future Work}

We presented PyKEEN-NSX, a modular extension of \textit{PyKEEN} that factors
negative sampling into a pool generator, conditioned on an explicit context,
and a selector. This abstraction unifies static, schema-aware and dynamic
strategies under a single interface, lets a new sampler be defined by its pool
alone, and turns pool size into a measurable property rather than an implicit
assumption. Using it, our preliminary analysis showed that constrained pools,
structural ones above all, rarely reach the requested $k$, and that under
random integration link prediction recovers the score of random corruption
itself: the encoded criterion is silently replaced by its fallback. The
extension makes this substitution controllable and observable without training.
Future work will extend the catalogue on the selector side, where score-based
strategies such as NSCaching and self-adversarial sampling remain
unimplemented, and broaden the evaluation across datasets and models.
Performance optimization, through caching and faster pool computation, and a
parallel implementation are also foreseen.

\section*{Declaration on Generative AI}
During the preparation of this work,  DeepL Write, Grammarly, and ChatGPT were used for grammar checking, rephrasing, and rewording.
After using these tools and services, the authors reviewed and edited the content as needed, taking full responsibility for the publication's content.

\bibliography{bib}

\end{document}